\documentclass[10pt,twocolumn,letterpaper]{article}

\usepackage{iccv}
\usepackage{times}
\usepackage{epsfig}
\usepackage{graphicx}
\usepackage{amsmath}
\usepackage{amssymb}
\usepackage{multirow}
\usepackage{multicol}
\usepackage{subcaption}
\usepackage[dvipsnames]{xcolor}
\usepackage{colortbl}
\usepackage{cite}
\usepackage{comment}
\usepackage{verbatim}
\usepackage{enumitem}
\usepackage{supertabular}
\usepackage[acronym]{glossaries}

\usepackage{siunitx} 
\usepackage{etoolbox} 

\makeglossaries

\usepackage[pagebackref=true,breaklinks=true,letterpaper=true,colorlinks,bookmarks=false]{hyperref}

 \iccvfinalcopy 

\def\iccvPaperID{1805} 

\definecolor{EPIC-COLOR}{HTML}{ED323E}
\definecolor{EPIC-COLOR2}{HTML}{00D6D6}

\definecolor{PT0}{RGB}{255,238,217}
\definecolor{PT1}{RGB}{255,236,213}
\definecolor{PT2}{RGB}{255,205,140}
\definecolor{PT5}{RGB}{201,87,0}
\definecolor{TL3}{RGB}{219,240,0}
\definecolor{TL4}{RGB}{201,231,0}
\definecolor{TD4}{RGB}{0,181,195}

\begin{document}

\title{Supervision Levels Scale (SLS)}

\author{Dima Damen and Michael Wray\\
University of Bristol, UK\\
\texttt{\{Dima.Damen, Michael.Wray\}@bristol.ac.uk}
}
\maketitle

\begin{abstract}
We propose a three-dimensional discrete and incremental scale to encode a method's level of supervision --- i.e. the data and labels used when training a model to achieve a given performance.
We capture three aspects of supervision, that are known to give methods an advantage while requiring additional costs: pre-training, training labels and training data.
The proposed three-dimensional scale can be included in result tables or leaderboards to handily compare methods not only by their performance, but also by the level of data supervision utilised by each method.
The Supervision Levels Scale (SLS) is first presented generally for any task/dataset/challenge. It is then applied to the EPIC-KITCHENS-100 dataset~\cite{Damen2020RESCALING}, to be used for the various leaderboards and challenges associated with this dataset.
\end{abstract}

\vspace*{-24pt}
\section{Motivation}
Performance gains for supervised learning methods have been achieved through a variety of means. 
Of these, additional loss/regularisation functions, architectural changes, and data augmentation techniques have all demonstrated high performance gains.
Acknowledging models are typically data-hungry, a primary way to improve performance is to increase the amount of labelled training data. However, this is restricted to problems where additional data is readily available and labelling costs are not an issue.
Alternatively, using a large-scale dataset for pre-training, before fine-tuning on the target data, has been shown to provide a significant advantage. 
Particularly of note, there has been a recent surge of methods using self-supervised pre-training which offers a competitive alternative to increasing the amount of labelled pre-training data.
Additionally, during fine-tuning, some methods use 
additional training labels, such as segmentation or object graphs, which have also provided performance boosts.

All of the above makes method comparisons far from trivial. 
One method can outperform another purely based on the amount and type of data used for pre-training (i.e. initialisation) or because additional labels have been utilised during training. 
We propose a discrete and incremental scale that can be incorporated into leaderboards for handily capturing the levels of supervision between methods, compared on the same test set.
Thus allowing direct comparison between methods that use the same level of supervision.
Additionally, this scale also enables assessing the impact of various dimensions of supervision on the performance.

\begin{figure}
    \centering
    \includegraphics[width=0.85\linewidth]{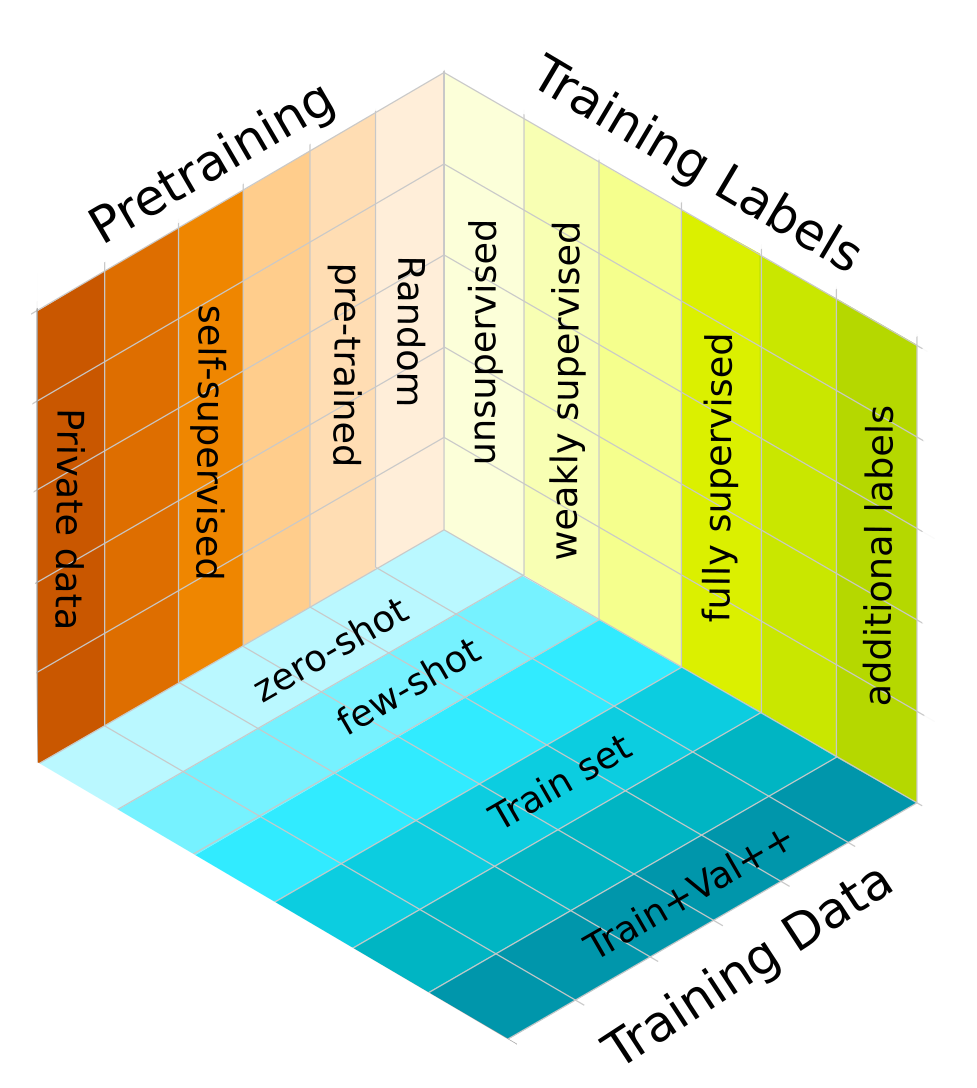}
    \caption{The three-dimensional SLS captures incrementally levels of supervision: pre-training, labels and amount of training data.}
    \label{fig:intro_figure}
\end{figure}

It is important to note that the proposed scale does not include any knowledge of the method's implementation details (e.g.~augmentation, optimisation choice, loss functions, ...). Instead, it focuses on the data-relevant knowledge available to the researchers before they make any method-relevant choices. 
This makes the scale suitable for anonymous submissions to leaderboards, not disclosing methods' novelties.

We thus restrict our scale to include three supervision dimensions (Fig~\ref{fig:intro_figure}). These are:
\begin{enumerate}
\vspace*{-4pt}
    \item Pre-training: The amount of data used in pre-training gives an advantage, particularly when the type of data used in pre-training is relevant to the task.
    \item Training labels: The nature of labels associated with the training data, is another dimension to distinguish between methods' supervision. For example, a method that takes advantage of not only image-level labels but also localisation or segmentation labels has an advantage for image captioning.
    \item Training data: The amount of training data offers a clear advantage. When less labelled data is utilised, while achieving comparable performance, such methods deserve to be highlighted. 
\end{enumerate}
\vspace*{-4pt}

In the next sections, we provide a general framework for this discrete three-dimensional scale, which we refer to as the Supervision Levels Scale (SLS). We believe SLS achieves the following objectives:
\begin{itemize}
    \item SLS can be used to readily compare methods in results tables, so as to highlight any data-relevant advantages when comparing state-of-the-art performances.
    \item SLS can be used in anonymous leaderboards in order to encourage and highlight methods that utilise less labels to demonstrate their competitive performance.
    \item SLS enables one leaderboard to compare a variety of self-supervised, weakly-supervised, few-shot and other distinct levels of supervision, as well as their combination. This would encourage the community to converge to a single evaluation leaderboard for the task
    , while acknowledging these methods differ in data usage.
    \item SLS can be used to analyse the impact of data-relevant advantages (pre-training, training data and training labels) for one task on the same test set.
\end{itemize}

We additionally demonstrate how SLS can be used in the leaderboards for the EPIC-KITCHENS dataset, showing how it can compare the supervision levels of previous years' competition winners.
This is achieved through self-declaration, i.e. each submission is requested to declare the supervision level across each dimension.

\section{Related Efforts}
The proposed Supervision Levels Scale (SLS) can be applied to provide additional supervision tags as methods are compared on the same test set.
Previous efforts to find discrete scales or categories have focused on identifying easy or hard examples within one datasets, in order to provide insights into methods failure cases (e.g.~\cite{Heoim_ECCV2012,Hosang2016,tide-eccv2020}). 
Up to our knowledge, there is no previous attempt to provide a discrete scale of the levels of supervision on the same dataset. 
We hope that this first attempt can be further utilised and elaborated for a variety of computer vision tasks.

\section{Supervision Levels Scale}
The proposed SLS has three dimensions: Pre-training~(PT), Training labels~(TL) and (amount of) Training data~(TD), explained below. These dimensions are orthogonal and a method is likely to consider a different level in each dimension. 
The discrete levels are incremental, with the lowest level implying the least amount of supervision on the corresponding scale.

For each dimension, we define five supervision levels (1-5), in addition to 0 where no supervision on that dimension is expected. 
A more fine-grained or wider range could be adopted by future tasks, but in this proposal we believe this level of discretisation to be sufficient.

\subsection*{SLS-PT: Pre-Training}
Pre-training is the data utilised to provide model initialisation, and is independent from data used in training the model itself. 
When the model is made of several parts or stages (e.g. feature extraction followed by temporal modelling), the PT supervision level is calculated for each stage, and the maximum PT value is considered as the supervision level of the overall method. For example, if a method uses features (pre-trained and then fine-tuned) while the classifier is trained from scratch, then the feature's pre-training is used to define the level of PT supervision of the full method.

\vspace{12pt}
\noindent\begin{supertabular}{|p{0.04\linewidth}|p{0.86\linewidth}|}
    \hline
    \textbf{PT} & \textbf{Description} \\ \hline
    0 & No pre-training was used. Models are randomly initialised, including features --- i.e. features are trained from randomly initialised models.\\
    1 & Standard pre-training on data of limited relevance to the task. For example, in a task on medical data, ImageNet~\cite{imagenet_cvpr09} is used for pre-training.\\
    2 & Relevant pre-training to the task. The data for pre-training is chosen to best fit the problem, and is believed to have a significant impact on learning low-level as well as medium-level features.\\
    3 & Self-supervision on large-scale unlabelled public data. Importantly, as public data is been used, the pre-training can be replicated.\\
    4 & Self-supervision using task-specific data. That is, a model has used training and/or test data on which performance will be trained/evaluated, with/without other large-scale public data
    , thus offering stronger pre-training supervision.\\
    5 & In addition to any or all of the above, pre-training on private data. This level of pre-training supervision is restricted to approaches that pre-train on data not accessible to other researchers and thus cannot be replicated. Even when these models are made available, other researchers are unable to replicate or improve this pre-training thus should be considered as an advantage.\\ \hline
\end{supertabular}

\subsection*{SLS-TL: Training Labels}
The second dimension of the SLS focuses on the training labels utilised by the method. This includes labels already available with the dataset, or additional labels acquired by the authors for the model specifically.

\vspace{12pt}
\noindent\begin{supertabular}{|p{0.04\linewidth}|p{0.86\linewidth}|}
    \hline
    \textbf{TL} & \textbf{Description} \\ \hline
    0 & No labelled supervision was used. Training data was used in an unsupervised or self-supervised manner without employing any labels.\\
    1 & Weak supervision - L1. Weak labels are provided for multiple instances. One-to-one mapping between labels and instances is not available.\\
    2 & Weak supervision - L2. Weak labels (noisy or incomplete) are associated with each instance.\\
    3 & Strong supervision - an instance-level label is given.\\
    4 & Strong supervision - in addition to instance-level labels, additional labels are provided (e.g. segmentation in images or spatio-temporal bounding boxes in video).\\
    5 & Strong supervision with additional labels, not available with the dataset by default.\\ \hline
\end{supertabular}

\subsection*{SLS-TD: Training Data}
Even with the same pre-training and labels supervision, methods can vary by the amount of training data used. The last dimension of the SLS captures these differences, unifying few-shot and many-shot training paradigms, when evaluated on the same test set.
Note that smart approaches to utilise less data (e.g.~avoiding noisy labels) can result in improved performance. However, in this case the data \emph{has been fully taken advantage of} and thus these methods have access to the same amount of training data.

\vspace{12pt}
\noindent\begin{tabular}{|p{0.04\linewidth}|p{0.86\linewidth}|}
    \hline
    \textbf{TD} & \textbf{Description} \\ \hline
    0 & Zero-shot learning - the training set has no label class/category overlap with the test set.\\
    1 & Few-shot learning - In line with~\cite{NIPS2016_6385}, (up to) 5-shot training data is used (per label class/category).\\
    2 & Efficient learning - Randomly selected fraction (commonly 25\%) of the data was used. The remainder of the training data were not used and the choice of sample is not optimised.\\
    3 & Train set - training set used in full.\\
    4 & Train+Val sets - after optimising any hyperparameters on the validation set, the combination of training and validation sets are used in training the model. Note that we here assume an official Train/Val split.\\
    5 & Train+Val++ sets - additional data is used during training (note that this is different from pre-training. This could be used with or without labels, but utilised during training the model itself.\\ \hline
\end{tabular}

\paragraph{Putting it together}
The three dimensions are then put together for each method, in the order: 
\begin{equation*}
    \text{\textbf{SLS-PT-TL-TD}}
\end{equation*}
To give an example, a method trained from scratch (i.e. with no pre-training), uses instance-level strong supervision as well as being trained with the full training set would be referred to as: \mbox{SLS-0-3-3}. Conversely, a method that uses private data for pre-training, weak-supervision from incomplete labels and few-shot learning would be referred to as: SLS-5-2-1. The two methods would be evaluated on the same test set, and thus the performance (given chosen evaluation metrics) are directly comparable, though with significantly different levels of supervision.

Next, we give an example of SLS on our dataset \mbox{EPIC-KITCHENS}~\cite{Damen2020RESCALING} and provide some comparative analysis from the public leaderboard of the 2019 and 2020 leaderboards of the Action Recognition challenge.

\begin{figure*}[t]
\includegraphics[width=\linewidth]{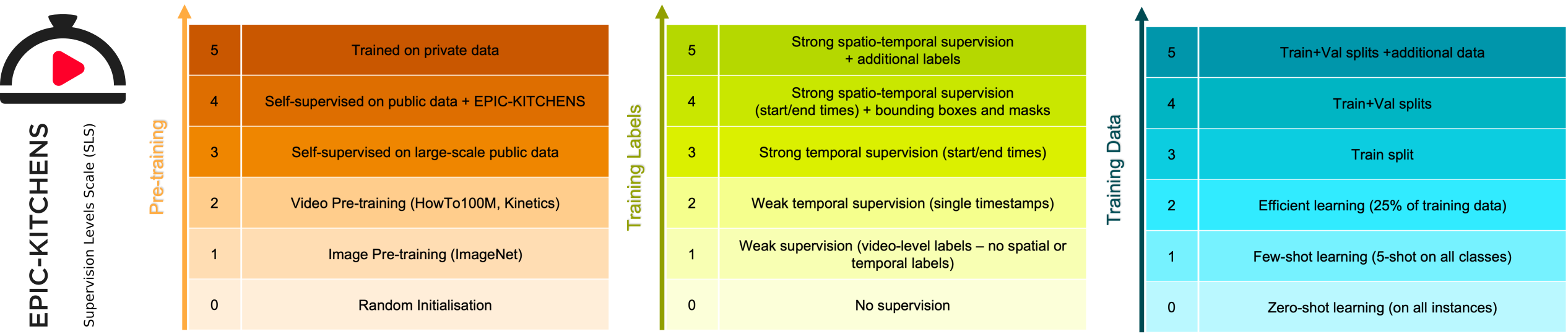}
\caption{SLS for the EPIC-KITCHENS Action Recognition challenge}
\label{fig:SLS-EPIC}
\end{figure*}

\begin{table*}[t]
\vspace{-1mm}
\begin{center}
\resizebox{\textwidth}{!}{%
\begin{tabular}{lcl|cl|ccc|SSS|SSS|}

                        & &        & \multicolumn{2}{c|}{\textbf{Submissions}} 
                         & \multicolumn{3}{c|}{\textbf{SLS}} 
                        & \multicolumn{3}{c|}{\textbf{Top-1 Accuracy}} & \multicolumn{3}{c|}{\textbf{Top-5 Accuracy}}\\\cline{4-14}
                        &\multicolumn{1}{c}{Rank}        &\multicolumn{1}{l|}{Team} &\multicolumn{1}{c}{Entries} &\multicolumn{1}{c|}{Date} & \multicolumn{1}{c}{PT}      & \multicolumn{1}{c}{TL}      & \multicolumn{1}{c|}{TD}    & \multicolumn{1}{c}{VERB}      & \multicolumn{1}{c}{NOUN}      & \multicolumn{1}{c|}{ACTION $\blacktriangle$}     & \multicolumn{1}{c}{VERB}   & \multicolumn{1}{c}{NOUN}  & \multicolumn{1}{c|}{ACTION}\\\hline
\multirow{10}{*}{\rotatebox{90}{\textbf{S1}}}
                        & 1 &  UTS-Baidu~\cite{Wang_2020_Symbiotic} &14 &05/28/20  &\cellcolor{PT5}5 &\cellcolor{TL4}4 &\cellcolor{TD4}4&  70.41 &  52.85 &  42.57 &  90.78 &  76.62 &  63.55 \\ 
                        & 2 &  NUS-CVML~\cite{Sener_2020_temporal} &18 &05/29/20  &\cellcolor{PT2}2 &\cellcolor{TL4}4 &\cellcolor{TD4}4&  63.23 &  46.45 &  41.59 &  87.50 &  70.49 &  64.11 \\
                        
\rowcolor{lightgray}
                        &  &  UTS-Baidu~\cite{Wang_2019_Baidu} &16 &05/30/19 &\cellcolor{PT2}{2} &\cellcolor{TL4}{4} &\cellcolor{TD4}{4} &  69.80 &  52.27 &  41.37 &  90.95 &  76.71 &  63.59 \\
                        
                         &3  &  SAIC-Cambridge~\cite{Perez_2020_Knowing} &34 &05/27/20 &\cellcolor{PT1}1 &\cellcolor{TL3}3 &\cellcolor{TD4}4&  69.43 &  49.71 &  40.00 &  91.23 &  73.18 &  60.53\\ 
                        
                         &3  &  FBK-HuPBA~\cite{Sudhakaran_2020_FBK} &50 &05/29/20 &\cellcolor{PT5}5 &\cellcolor{TL3}3 &\cellcolor{TD4}4&  68.68 &  49.35 &  40.00 &  90.97 &  72.45 &  60.23\\ 
                         
                         &4  & GT-WISC-MPI~\cite{liu2019forecasting} &12 &01/30/20 &\cellcolor{PT2}2 &\cellcolor{TL3}3 &\cellcolor{TD4}4&68.51	&49.96	&38.75	&89.33 &72.30 &58.99 \\
                         
                         &5  & G-Blend~\cite{Wang_2020_what} &14 &05/28/20 &\cellcolor{PT5}5 &\cellcolor{TL3}3 &\cellcolor{TD4}4 &66.67	&48.48	&37.12	&88.90 &71.36 &56.21\\
\rowcolor{lightgray}
                        &6  &  TBN~\cite{Kazakos_2019_ICCV} &2 &05/30/19&\cellcolor{PT2}2 &\cellcolor{TL3}3 &\cellcolor{TD4}4&  66.10 &  47.89 &  36.66 &  91.28 &  72.80 &  58.62\\
\rowcolor{lightgray}
                        & &  FAIR~\cite{Ghadiyaram_2019_CVPR} &9 &10/30/19&\cellcolor{PT5}5 &\cellcolor{TL3}3 &\cellcolor{TD4}4&  64.14 &  47.65 &  35.75 &  87.64 &  70.66 &  54.65\\
 
\end{tabular}}
\vspace*{-8pt}
\caption{Top Teams in the 2019 (gray) and 2020 Action Recognition Challenges in EPIC-KITCHENS-55. Coloured columns include the added SLS demonstrating the various levels of supervision employed by these methods.}
\label{tab:resultsAR}
\end{center}
\end{table*}

\begin{figure*}
\includegraphics[width=\linewidth]{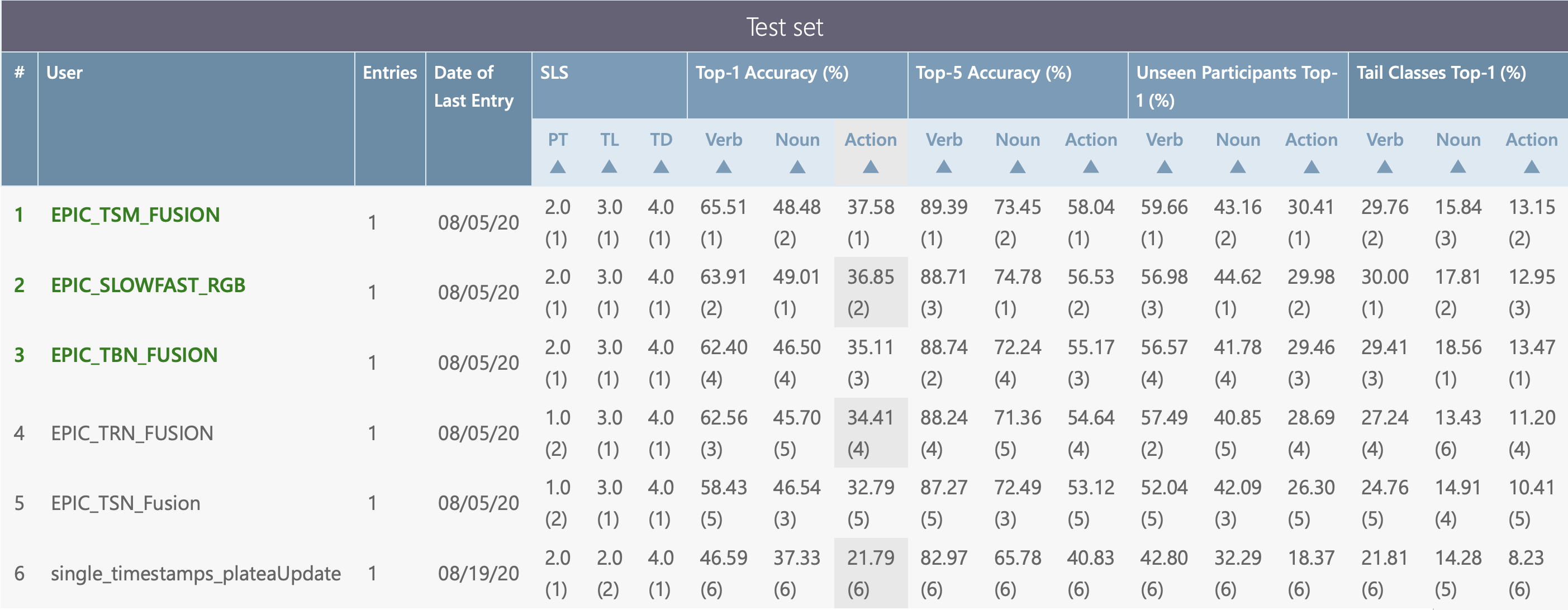}
\caption{Recently released EPIC-KITCHENS-100 Action Recognition Leaderboard, with self-reporting submission of SLS on the various baselines. Note 6 only uses weak supervision hence the SLS-TL of 2.}
\label{fig:epic-new-leaderboard}
\end{figure*}

\section{SLS for EPIC-KITCHENS}

EPIC-KITCHENS is the largest egocentric dataset, captured in a non-scripted manner, and recorded in participants' kitchens. 
First released in 2018~\cite{Damen2018EPICKITCHENS}, and later extended~\cite{Damen2020RESCALING} to 20M frames, 90K action segments and 100~hours of densely annotated actions, the dataset offers an ideal testbed for assessing video understanding benchmarks in untrimmed videos. 
Several benchmarks are associated with the dataset - currently 5 leaderboards are available for comparative evaluation, see challenges at: 

\noindent{\url{https://epic-kitchens.github.io/}}

We showcase how SLS can apply to the \mbox{EPIC-KITCHENS} Action Recognition (AR) challenge (Sec~4.1 in~\cite{Damen2020RESCALING}), then demonstrate how it can be used to compare the levels of supervision of methods submitted to this leaderboard.

We specialise the SLS for this task as follows:
\begin{itemize}
\setlength\itemsep{0em}
    \item SLS-PT for AR in EPIC-KITCHENS:
    \begin{itemize}
        \item SLS-1-X-X: standard pre-training, in this case would be any image-based pre-training.
        \item SLS-2-X-X: Relevant pre-training would be video pretraining (e.g. large scale video datasets such as Kinetics~\cite{Carreira_2017_CVPR} or HowTo100M~\cite{miech19howto100m}).
    \end{itemize}
    \item SLS-TL for AR in EPIC-KITCHENS:
    \begin{itemize}
        \item SLS-X-1-X: Would be utilising no temporal information about action segments or instances per video, i.e. video-level labels of actions are only known, without their rough or exact locations.
        \item SLS-X-2-X: Would utilise weak temporal labels, known as single timestamps in EPIC-KITCHENS~\cite{moltisanti19action}. These are rough single time points within or close to the action of interest.
        \item SLS-X-3-X: Full temporal labels (i.e. start-end times) without any spatial labels have been used.
        \item SLS-X-4-X: Methods have taken advantage of spatio-temporal labels, i.e. start-end times plus bounding boxes, hand detections and/or masks.
    \end{itemize}
    \item SLS-TD for AR in EPIC-KITCHENS: No adjustments needed to adapt this dimension. Train and Val splits correspond to the splits in Table~1 from~\cite{Damen2020RESCALING}.
\end{itemize}

Fig.~\ref{fig:SLS-EPIC} visualises SLS for AR in EPIC-KITCHENS, across the three dimensions. We next demonstrate how this can be applied to the leaderboard, using examples from the previous years' challenge leaderboards. Note that these were evaluated on a previous version (subset) of the dataset known as EPIC-KITCHENS-55.

In Table~\ref{tab:resultsAR} we list the winners from the AR challenge for years 2019 and 2020 ranked by their performance. More details of these methods are available in the technical reports~\cite{Damen_2019_Challenges, Damen_2020_Challenges}. We denote SLS for each method based on the technical reports submitted. The table shows that: methods vary in pre-training, with some utilising the weights from the private dataset IG-Kinetics~\cite{Ghadiyaram_2019_CVPR}. Methods that employed spatio-temporal labels (TL=4) as opposed to temporal labels solely, consistently outperform others, highlighting the nature of this dataset: actions take place  in the messy environment of participants' kitchens and identifying where the action takes place is consistently helpful. Finally, no work has attempted few-show or efficient learning on the dataset to-date. SLS-TD in this table is thus consistent for all methods.

We have also integrated SLS into the new leaderboard for Action Recognition in EPIC-KITCHENS-100, open for the 2021 round of the challenges. Fig~\ref{fig:epic-new-leaderboard} shows the leaderboard recently opened with the baselines from~\cite{Damen2020RESCALING}. Importantly, this leaderboard shows both full temporal supervised (TL=3) as well as weak temporal supervision (TL=2) in the same leaderboard. This enables direct comparison of the drop in performance between methods that utilise single timestamps~\cite{moltisanti19action} and full temporal supervision. In the given baseline, start-end times improve Top-1 Action Accuracy by 11-16\%.

\section{Further discussion and limitations of SLS}
There are a few known limitations of the proposed SLS, due to decisions that were made with the aim to reduce complexity.
First, the discrete nature of SLS only indicates part of the picture, and details to replicate any method are only available by reading about the methods themselves. We believe this compromise to be acceptable.
Second, when a method is formed of multiple stages (e.g. feature extraction then temporal modelling for example), SLS encodes only one scale per dimension; the maximum level of supervision across all the stages. Due to some models using a large number of stages and sources for pre-training, we also consider this simplicity to be acceptable.
Third, SLS relies on self-reporting of each method's supervision scale, and it is left to authors to justify their declared supervision levels. Misunderstanding any dimension can result in incorrect reporting. We hope that this technical report can assist in alleviating any ambiguity, and will be reviewing its content as we receive inquiries or concerns.

The remaining (fourth and fifth) limitations require future work to be addressed.
Fourth, SLS assumes all training data uses the same training labels. Semi-supervised approaches cannot be correctly represented by SLS.  
Fifth, multi-modal data is not captured. For example, in Table~\ref{tab:resultsAR} the method TBN~\cite{Kazakos_2019_ICCV} utilises audio information in addition to video, giving it an advantage. Similarly, when performing tasks like retrieval or captioning, two modalities (vision + langauage) will consider different SLS for each modality. We leave this consideration of multiple modalities and their relationship to SLS for future work and focus only on the visual modality.

\section{Conclusion}
This report introduces a three-dimensional, incremental and discrete scale to encode the level of supervision of methods, compared on the same test set. It aims to unify and directly compare methods that attempt self-supervised pre-training, few-shot learning and weak supervision, when compared for the same task and on one test set. 
We believe SLS can provide new insights into how methods learn from more/less labelled data.
We will be analysing the SLS of various methods submitted to the five leaderboards of EPIC-KITCHENS-100 in the next round of challenge submissions.


\paragraph{Acknowledgement.} We would like to thank Hazel Doughty and Will Price for early discussions on this idea.
\vspace*{-4pt}
{\small
\bibliographystyle{ieee}
\bibliography{egbib}
}

%

\end{document}